\documentclass[superscriptaddress,twocolumn,prx]{revtex4-1}
\usepackage{graphicx}
\usepackage{amsmath,amssymb,mathrsfs}
\usepackage{natbib}
\usepackage{subfigure}
\usepackage{tabularx}
\usepackage{epsfig}
\usepackage{longtable}
\usepackage{amsfonts}
\usepackage{rotating}
\usepackage{bbold}
\usepackage{hhline}
\usepackage{braket}
\usepackage{txfonts, comment}
\usepackage{multirow}
\usepackage{appendix}
\usepackage[unicode=true,bookmarks=true,bookmarksnumbered=false,
bookmarksopen=false,breaklinks=false,pdfborder={0 0 1},backref=false,colorlinks=true]{hyperref}
\hypersetup{linkcolor=magenta,urlcolor=blue,citecolor=blue,
pdfstartview={FitH},hyperfootnotes=false,unicode=true}

\def\be{\begin{equation}}
\def\ee{\end{equation}}
\def\bea{\begin{eqnarray}}
\def\eea{\end{eqnarray}}

\begin{document}

\title{Impact of ChatGPT on the writing style of condensed matter physicists}

\author{Shaojun Xu}
\affiliation{School of Economics and Management, Zhejiang Sci-Tech University, Hangzhou 310018, China}
\email{sjxu\_2010@126.com}

\author{Xiaohui Ye}
\affiliation{School of Economics and Management, Zhejiang Sci-Tech University, Hangzhou 310018, China}

\author{Mengqi Zhang}
\affiliation{School of Economics and Management, Zhejiang Sci-Tech University, Hangzhou 310018, China}

\author{Pei Wang}
\affiliation{Department of Physics, Zhejiang Normal University, Jinhua 321004, China}
\email{wangpei@zjnu.cn}

\date{\today}

\begin{abstract}
We apply a state-of-the-art difference-in-differences approach 
to estimate the impact of ChatGPT’s release on the writing style 
of condensed matter papers on arXiv. Our analysis reveals a 
statistically significant improvement in the English quality of abstracts 
written by non-native English speakers. Importantly, this improvement 
remains robust even after accounting for other potential factors, 
confirming that it can be attributed to the release of ChatGPT. 
This indicates widespread adoption of the tool. Following the release of ChatGPT,
there is a significant increase in the use of unique words, while the 
frequency of rare words decreases. Across language families, 
the changes in writing style are significant for authors from the Latin 
and Ural-Altaic groups, but not for those from the Germanic or other Indo-European groups.
\end{abstract}

\maketitle

%\date{\today}

\section{Introduction}

Since its release, ChatGPT~\cite{Kim2023},
a popular large language model (LLM), has been widely applied
across various fields, including healthcare~\cite{Jonathan2023,Gupta2023, 
Ali2023, Janssen2023, Patel2023, Howard2023, Hirosawa2023},
finance~\cite{Li2023,Fatouros2023}, social media~\cite{Ye2023, Srivastava2023},
education~\cite{Nunes2023,Joshi2023,Choi2023, OConnor2023}, 
medical licensing and bar exams~\cite{Gilson2023,
Huh2023,Bommarito2022,Chalkidis2023}, coding~\cite{Savelka2023},
and scientific research~\cite{Kantor2023}.
Specifically, in academic writing, ChatGPT can generate 
seemingly genuine scientific papers and peer-review reports~\cite{Stokel2022,Else2023},
sparking significant discussion about its 
applications~\cite{Stokel2023,OConnor2023,Verhoeven2023}.
ChatGPT and other LLMs act as double-edged swords 
for the scientific community~\cite{Shen2023}.
While concerns about potential threats such as reduced 
rigor, plagiarism, or copyright issues persist,
some journals~\cite{Thorp23} have banned the use of ChatGPT 
or similar tools. However, ChatGPT also helps reduce linguistic 
discrimination in the publication process by partially bridging 
the gap between native and non-native English speakers, 
thereby saving time on writing and speeding up scientific communication.

Beyond qualitative discussions, quantitative studies have 
investigated whether the scientific community is genuinely 
using ChatGPT for academic writing. Recently, researchers 
have estimated the utilization rate of ChatGPT based on statistical frameworks~\cite{Astarita24,Geng24,Liu24,Liang2404,Liang2403,Gray24}.
These studies analyze the frequency of specific 
keywords~\cite{Liang2403,Liang2404} that differentiate between 
texts generated by ChatGPT and those written by human researchers.
The large-scale analysis indicates a sharp increase in AI-modified 
content after the ChatGPT release, with the utilization rate varying across disciplines.
Computer scientists appear to be the most enthusiastic adopters~\cite{Liang2404},
with approximately 35\% of abstracts revised using ChatGPT~\cite{Geng24}.
A widespread adoption of ChatGPT in the astronomy papers
is also observed~\cite{Astarita24}.

Previous studies have distinguished ChatGPT-modified papers from 
non-ChatGPT papers by counting keywords frequencies. 
In this paper, we employ a state-of-the-art 
method based on policy effect evaluation in econometrics, i.e. 
difference-in-differences (DID) method, to investigate the 
impact of ChatGPT's release on the writing style of scientific papers. 
The DID technique, originally developed by Ashenfelter and Card~\cite{Ashenfelter1985} 
to estimate causal relationships in non-experimental settings where 
random assignment is not possible, accounts for time-invariant 
unobserved heterogeneity between a treatment group and a 
control group by comparing the changes in outcomes over time. 
This allows us to isolate the effect of ChatGPT's release. Here, 
we use native English speakers as the control group, previously 
found to be less enthusiastic about using ChatGPT~\cite{Liu24}, 
and non-English speakers as the treatment group, classifying 
papers based on the author's native language families. This 
approach enables us to determine whether ChatGPT has 
been significantly adopted in the writing process.

Previous studies primarily relied on machine learning and 
other large language models to identify keywords for
determining whether a particular paper was written using 
ChatGPT. In contrast, we employed the “Grammarly” software 
to quantify the English writing style, thereby assessing whether 
the release of ChatGPT has led to improvements in English 
writing quality. 
By using Welch's t-test, the paired t-test, and the DID 
method, this study comprehensively demonstrates the impact of 
ChatGPT’s release on multiple dimensions, including overall 
writing quality, readability, word usage characteristics, and 
grammatical errors, all of which characterize a paper’s writing style. 
% Welch's t-test also known as unequal variances t-test is used when you want to test whether the means of two population are equal. 

Our results support the hypothesis that after ChatGPT's release, 
the writing quality of condensed matter physics papers by non-native 
English authors has significantly improved, indicating widespread 
adoption of ChatGPT within this community. 
Furthermore, ChatGPT not only enhances readability but also reduces grammatical errors. 
After its release, there is a notable increase in the 
use of unique words and a decrease in rare words.
Comparisons among different language families reveal 
that English writing quality has significantly improved for authors from the Latin and 
Ural-Altaic groups , while no such 
effect is observed in the Germanic and other Indo-Euro language groups. 
This suggests that ChatGPT 
adoption varies depending on the author's language family.
Our quantitative findings provide a basis for the scientific community, 
including authors, editors, and referees, to consider how 
to treat the use of ChatGPT in scientific writing and 
potentially establish community-wide standards in the future.

The rest of the paper is organized as follows.
The dataset is described in Sec.~\ref{sec:method}. The results of Welch's t-test and paired t-test are presented in Sec.~\ref{sec:element}. 
The empirical results from DID analysis are discussed in Sec.~\ref{sec:ecoana}.
Finally, Sec.~\ref{sec:summ} is a short summary.

\section{Description of dataset}
\label{sec:method}

We choose arXiv as the primary source for papers, following the
approach of previous researchers~\cite{Liang2404,Geng24,Liu24}.
ArXiv is particularly well-suited for DID analysis, in which comparing
the writing date of paper with the release date of ChatGPT is crucial. 
In fields such as physics, mathematics, and computer science, 
most papers are uploaded to arXiv before journal submission.
A key advantage of arXiv is that papers appear on the platform 
immediately after completion, making their writing dates easily 
accessible. In contrast, journal publication often occurs long after a 
paper is written due to the review process. Furthermore, some papers 
undergo multiple rounds of submission and rejection, making their 
writing dates difficult to trace through journal websites.

We focus on two subcategories of condensed matter 
physics on arXiv: "Statistical Mechanics" (cond-mat.stat-mech) 
and "Disordered Systems and Neural Networks" (cond-mat.dis-nn). 
These subcategories are closely related to statistics, computer science and 
artificial intelligence, both in terms of methods and models, 
increasing the likelihood of ChatGPT-assisted copyediting~\cite{Liang2404,Geng24}.
We analyze the abstract of each paper to assess its writing style. 
While some previous studies have examined full texts, others 
have focused exclusively on abstracts for writing style analysis~\cite{Geng24,Liu24}.
In this study, we focus solely on abstracts as a preliminary step in applying the DID approach. 
Since ChatGPT was released on November 30, 2022, we selected 
papers from arXiv that appeared between January 1, 2020, and 
May 31, 2023, a period that spans the release of ChatGPT.

A pairing process is essential to our approach, and it is absent 
in previous studies. The concept of pairing is based on the assumption 
that an individual author will change their writing style after 
adopting ChatGPT assistance. Therefore, comparing papers by 
the same author before and after the release of ChatGPT can 
help determine whether ChatGPT has been used.
The pairing process is as follows: First, we identify authors who 
have written at least one paper after the release of ChatGPT, 
specifically between December 1, 2022, and May 31, 2023. Next, 
we search for papers written by the same authors during the period 
from January 1, 2020, to November 30, 2022. This pairing method 
allows us to systematically eliminate changes in writing style that 
could result from different authors contributing before and after the release of ChatGPT.
It is important to note that we choose first authors as the 
primary contributors to a paper's style, as they are typically 
responsible for drafting the manuscript in cases of multiple authors.

\begin{figure}[tbp]
\includegraphics[width=0.9\linewidth]{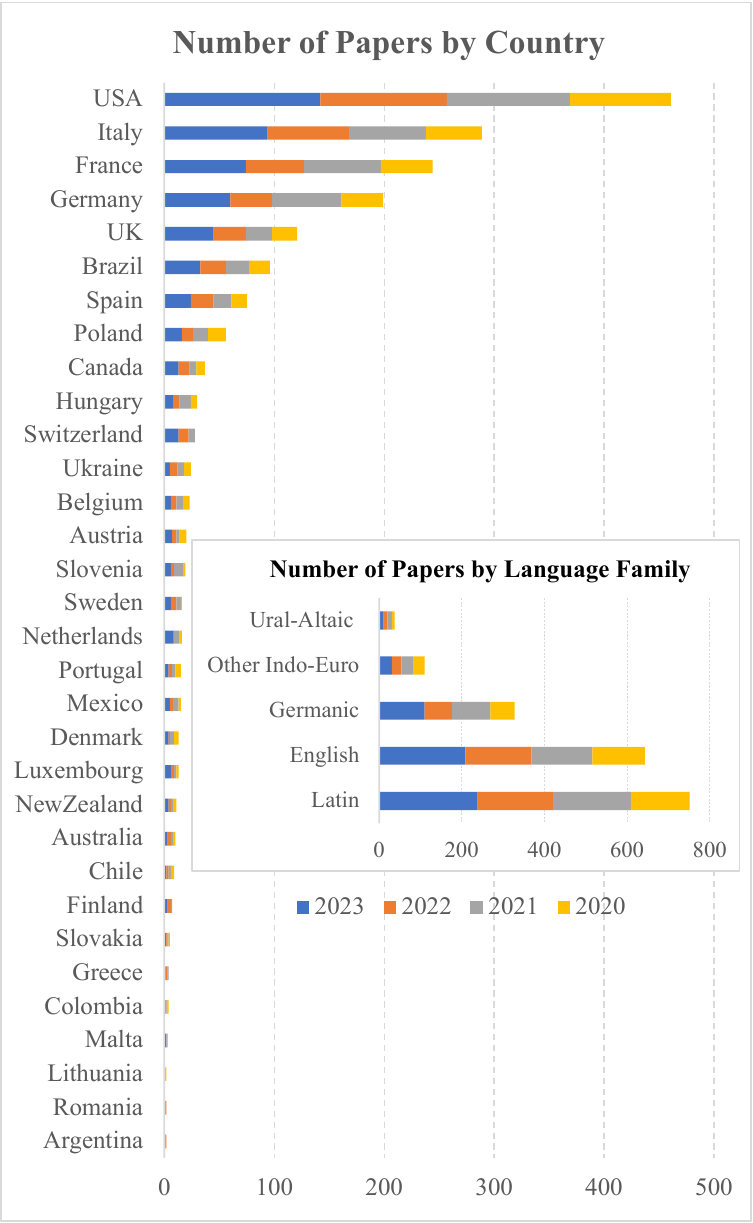}
\caption{Number of papers by country and language family.
Here, the Latin language family includes French, Italian, Spanish, Portuguese, and Romanian. 
The Germanic language family (excluding English) comprises German, Dutch, Luxembourgish, Danish, and Swedish. 
The other Indo-Euro language family includes branches of the Indo-European language family other than the Germanic and Latin branches, 
specifically the Slavic, Greek, and Armenian branches (e.g., Ukrainian, Polish, Slovak, Slovenian, Greek, Lithuanian, and others). 
The Ural-Altaic language family includes Finnish and Hungarian.
 }\label{fig1}
\end{figure}

In this paper, we exclude authors from Russia, Venezuela, and 
Asian countries, where formal access to ChatGPT is not guaranteed,
and where the language families are quite different from the Indo-Euro language family.
% We hope future studies will include these countries.
We analyze the influence of authors' native language families 
on the adoption of ChatGPT. To do this, we need to identify an author's 
native language. While the most accurate method would be to 
check their place of birth, obtaining detailed biographical information 
for some authors can be challenging. Therefore, we use the country 
of the author's affiliated institution as a proxy. For simplicity, we 
continue to use the term "native language".
If an author's affiliated institution changes during the period 
of interest, we use their final affiliation to identify their "native language". 
Our sample check shows that out of 501 authors, only 30 moved 
to institutions belonging to a different language family.
We categorize the authors into English and non-English groups. 
The non-English group is further divided into the Indo-Euro
language family, which includes the Germanic, Latin, and other branches, 
and the Ural-Altaic family. It's worth noting that languages in the 
Germanic branch are closest to English, as English itself belongs 
to the Germanic family (see Fig.~\ref{fig1} for the number of sample 
papers by country and language family).

To quantify the writing style of an abstract, we use "Grammarly" 
(https://www.grammarly.com/) to evaluate it across several dimensions, 
including the number of characters (denoted as $characters$), 
number of words ($words$), number of sentences 
($sentences$), average word length ($word\_len$), 
and average sentence length ($sentence\_len$). More important, the overall 
score of an abstract ($score \in [0,100]$) assesses its English quality, making 
$score$ our primary dependent variable. A higher $score$ indicates a better-quality abstract.
The second key variable is $read\_score$, which measures the 
readability of the abstract based on Flesch reading ease test. 
"Grammarly" also tracks the number 
of writing issues or errors ($issues$), the percentage of unique 
words ($unique$) to assess vocabulary diversity (i.e., the 
proportion of words used only once), and the percentage of rare
words ($rare$), which reflects the depth of vocabulary based on words 
that are not among the 5000 most common in English.

Additionally, we collect the following paper properties: whether 
there are any native English-speaking authors besides the first 
author ($others\_Eng = 1$ if there are, $others\_Eng = 0$ if there 
are none), the total number of authors ($authors$), and the 
subcategory of the paper ($stat=1$ for Statistical Mechanics, and $stat=0$ otherwise; 
$dis=1$ for Disordered Systems and Neural Networks, and $dis=0$ otherwise). 
We also track whether the paper has been published ($if\_pub=1$ 
for published, $if\_pub=0$ for not yet published) as of the data 
collection date. Note that paper information was manually collected 
between June~1 and July~18, 2023.

To eliminate the influence of extreme samples, we exclude the 
top and bottom 1\% of papers based on their $score$. After 
this adjustment, a total of 1,869 papers authored by 501 individuals 
from 32 countries remain for analysis.
As shown in Fig.~\ref{fig1}, authors from the USA produce the 
most papers (461), followed by Italian authors with 289 papers. 
Following the release of ChatGPT, the number of papers 
increases to 600, compared to 439 papers in the year prior (2022), 
and 469 and 361 papers in 2021 and 2020, respectively.
The inset of Fig.~\ref{fig1} highlights that the Latin subgroup 
produces the highest number of papers, with 194 authors contributing 751 papers. 
The English-speaking group ranks second, with 170 authors producing 643 papers. 
Next, the Germanic subgroup (excluding English) has 101 authors who write 328 papers,
while the other Indo-Euro subgroup contributes 110 papers from 27 authors.
Finally, the Ural-Altaic subgroup produces 37 papers from 9 authors.

Table~\ref{Tab:Basic} provides a summary of the descriptive
statistics for the variables. The average $score$ for the 1,869 abstracts is 87.01, 
with a standard deviation of 8.30. The $score$ ranges from a 
minimum of 60.00 to a maximum of 99.00. The average readability
score ($read\_score$) is 17.56, with a standard deviation of 12.46. 
On average, there are $62\%$ unique words and $36\%$ rare words in an abstract.
The average number of writing issues is 8.40. The average number of authors per paper 
is 2.87, while there are 35\% papers that have native English-speaking coauthors.
Each abstract contains an average of 1,070.57 characters, 
158 words, and 6.31 sentences. The average word length is 5.66 
characters, and the average sentence length is 25.67 words.

\begin{table}[tbp]
\renewcommand\arraystretch{2.0}
\resizebox{1.0\linewidth}{!}{
\begin{tabular}{| c  | c | c |  c | c | c | c | c |}
\hline
       & {\bf Mean} & {\bf S. D.}  & {\bf Minimum} & {\bf Median} & {\bf Maximum} &
             {\bf Skew} & {\bf Kurt} \\
\hline
${score}$       & 87.01 & 8.30  & 60.00 & 88.00 & 99.00 & -0.77 & 0.26 \\
\hline
${read\_score}$ & 17.56 & 12.46 & -41.00 & 18.00 & 77.00 & -0.16 & 0.45 \\
\hline
${unique}$ & 0.62 & 0.07 & 0.41 & 0.62 & 0.94 & 0.29 & 0.29 \\
\hline
${rare}$ & 0.36 & 0.05 & 0.18 & 0.36 & 0.56 & 0.06 & 0.10 \\
\hline
${issues}$ & 8.40 & 4.32 & 1.00 & 8.00 & 26.00 & 0.77 & 0.59 \\
\hline
${others\_Eng}$ & 0.35 & 0.48 & 0.00 & 0.00 & 1.00 & 0.64 & -1.59 \\
\hline
${authors}$ & 2.87 & 1.57 & 1.00 & 3.00 & 14.00 & 1.81 & 7.20 \\
\hline
${stat}$ & 0.86 & 0.35 & 0.00 & 1.00 & 1.00 & -2.09 & 2.36 \\
\hline
${dis}$ & 0.25 & 0.44 & 0.00 & 0.00 & 1.00 & 1.13 & -0.72 \\
\hline
${if\_pub}$ & 0.61 & 0.49 & 0.00 & 1.00 & 1.00 & -0.46 & -1.79 \\
\hline
${characters}$ & 1070.57 & 359.81 & 113.00 & 1034.00 & 2335.00 & 0.39 & -0.35 \\
\hline
${words}$ & 158.00 & 53.94 & 31.00 & 150.00 & 350.00 & 0.48 & -0.23 \\
\hline
${sentences}$ & 6.31 & 2.21 & 1.00 & 6.00 & 15.00 & 0.65 & 0.45 \\
\hline
${word\_len}$ & 5.66 & 0.39 & 4.00 & 5.70 & 7.00 & 0.15 & 0.19 \\
\hline
${sentence\_len}$ & 25.67 & 5.73 & 4.90 & 24.90 & 84.70 & 1.73 & 9.72 \\
\hline
\end{tabular}
}
\caption{Descriptive statistics of the variables. 
Here, S.~D. is the abbreviation of standard deviation.}\label{Tab:Basic}
\end{table}

\section{Welch's t-test and Paired t-test}
\label{sec:element}

As an appetizer, we first use Welch's t-test~\cite{Welch1947} 
to assess whether there is a statistically significant 
difference in writing quality (measured by $score$) 
before and after the release of ChatGPT. Note that
Welch's t-test offers a robust alternative to the
standard Student's t-test, accommodating unequal variances or sample 
sizes between the two sets, and has been widely applied 
across various fields, including healthcare, medicine~\cite{Ransohoff1978}, 
psychology, social sciences~\cite{Baron1986}, and economics and business~\cite{Card1994}.

The results are presented in Tab.~\ref{Tab:Welch}. 
The average $score$ after the release of ChatGPT is 87.69, compared to 
86.69 before its release, with a difference of 1.00. More important, 
this difference is statistically significant at the 5\% level, with a corresponding 
t-value of 2.49, indicating that writing skills have significantly improved after the release of ChatGPT.

Additionally, we compare the $score$ differences across specific years in 
Tab.~\ref{Tab:Welch}. We find that the $score$ in 2023 is significantly 
higher than in 2021 (86.46) and 2020 (86.57), with t-values of the difference being 2.38 and 2.06, 
respectively. However, the differences in $score$ among 2022, 2021, 
and 2020 are not statistically significant. For example, the t-value of the difference between 
2022 and 2021 is 1.04, which is not significant even at the 10\% level. 
Similarly, the difference between 2021 and 2020 is insignificant, with a t-value of -0.18.

\begin{table}[tbp]
\renewcommand\arraystretch{2.0}
\resizebox{0.9\linewidth}{!}{
\begin{tabular}{| c | c | c | c | c |}
\hline
             & {\bf Mean (1)} & {\bf Mean (2)}  & {\bf t-value} & {\bf df} \\
\hline
$\text{ (1)After - (2)Before}$ & 87.69 & 86.69 & $2.49^{**}$ & 1243.00 \\
\hline
$\text{ (1)Year2023 - (2)Year2022}$ & 87.69 & 87.05 & 1.27 & 927.56 \\
\hline
$\text{ (1)Year2023 - (2)Year2021}$ & 87.69 & 86.46 & $2.38^{**}$ & 953.66 \\
\hline
$\text{ (1)Year2023 - (2)Year2020}$ & 87.69 & 86.57 & $2.06^{**}$ & 782.57 \\
\hline
$\text{ (1)Year2022 - (2)Year2021}$ & 87.05 & 86.46 & 1.04 & 905.99 \\
\hline
$\text{ (1)Year2022 - (2)Year2020}$ & 87.05 & 86.57 & 0.82 & 762.26 \\
\hline
$\text{ (1)Year2021 - (2)Year2020}$ & 86.46 & 86.57 & -0.18 & 791.87 \\
\hline
\end{tabular}
}
\caption{Welch's t-test on the variable $score$. "df" denotes the degree of freedom. 
The superscript $**$ indicates that the difference is significant at the 5\% level.
"Year2023" represents the time period from December 2022 to May
 2023 (after the release of ChatGPT). "Year2022" represents the 
 period from January 2022 to November 2022
 (before the release of ChatGPT). "Year2021" and "Year2020"
represents the years of 2021 and 2020, respectively. }\label{Tab:Welch}
\end{table}

Compared to Welch’s t-test, the paired t-test is more suitable for
“before-and-after” comparison studies, because our data 
consists of matched pairs. 
In a paired t-test, the differences between 
each pair of observations (post-ChatGPT release minus pre-release) are calculated to determine 
whether the mean of these differences is significantly different from zero. 
This approach provides a clearer assessment of the treatment effect by 
effectively controlling for confounding variables that might affect the 
$score$. 
Therefore, we employ the paired t-test to estimate the 
impact of ChatGPT's release on writing style by comparing the $score$ 
of the same author before and after its release.

We calculate the average $score$ of an author's abstracts before 
the release of ChatGPT as the "before" value and the average $score$ 
of their abstracts after the release as the "after" value, constructing 
a paired "before-and-after" dependent variable for each author. 
Authors are further categorized into different language 
families, as described in Sec.~\ref{sec:method}.
The paired t-test results are shown in Tab.~\ref{Tab:Paired}.

Among the total 1,869 papers by 501 authors, the average $score$ for each author
after ChatGPT's release is 87.72, compared to 86.63 before its release, 
yielding a difference of 1.09. The $score$ after the release of ChatGPT
is significantly higher than before at the 5\% confidence level, with a corresponding t-value of 2.54.

Next, we assess the significance of this difference across authors' language families. 
In the English group (643 papers by 170 authors), the "before-and-after" 
paired $score$ difference is not significant, with a difference of only 0.35. 
This suggests that for native English speakers, ChatGPT's release did 
not significantly enhance their English writing skills, or they did not widely 
adopt ChatGPT for assistance. This finding aligns with previous 
research~\cite{Liu24}, supporting our use of native English speakers as 
a control group in the subsequent DID analysis.

However, in the Non-English group (1,226 papers by 331 authors), the 
"before-and-after" paired $score$ difference is 1.47, which is significant at 
the 5\% level, with a t-value of 2.73. This indicates that the $score$ significantly improved after the release of ChatGPT compared to before.
We further analyze these differences 
across various language families to explore how the impact varies.

For the Latin group (including French, Italian, Spanish, 
Portuguese, and Romanian), we find a significantly higher $score$ 
after ChatGPT's release, with a difference of 1.44 compared to before. 
This difference is significant at the 5\% level, with a t-value of 2.10. 
For the Germanic group (including German, Dutch, Luxembourgish, 
Danish, and Swedish), we find no significant difference before and 
after ChatGPT's release, with a difference of 1.28 and a t-value of 1.32. 
This result is expected, given that English is a Western Germanic language, 
making it less likely for native speakers of other Germanic languages to rely on ChatGPT for writing.

In the other Indo-Euro group, which includes the Slavic, Greek, 
and Armenian language families (such as Ukrainian, Polish, Slovak, Slovenian, Greek, Lithuanian, and others), 
we also find no significant differences before and after 
ChatGPT's release, with a t-value of 0.29. The average $score$ after 
ChatGPT's release is 85.20, compared to 84.52 before, representing 
a relatively small difference among all language groups. 
In contrast, the Ural-Altaic group (including Finnish and Hungarian). 
shows a significantly 
improvement in $score$ after ChatGPT's release, with a difference of 6.50. 

Thus, the Latin and Ural-Altaic groups exhibit significant improvements 
in English writing levels following ChatGPT's release, indicating a 
notable adoption of ChatGPT in scientific writing.

\begin{table}[tbp]
\renewcommand\arraystretch{2.0}
\resizebox{1.0\linewidth}{!}{
\begin{tabular}{| c | c | c | c | c | c |}
\hline
             & {\bf Mean (After)} & {\bf Mean (Before)}  & {\bf Difference} & {\bf t-value} & {\bf Observation}\\
\hline
$\text{All}$ & 87.72 & 86.63 & 1.09 & $2.54^{**}$ & 1869 (501) \\
\hline
$\text{English}$ & 88.25 & 87.90 & 0.35 & 0.50 & 643 (170) \\
\hline
$\text{Non-English}$ & 87.45 & 85.98 & 1.47 & $2.73^{**}$ & 1226 (331) \\
\hline
$\text{Latin}$ & 87.35 & 85.90 & 1.44 & $2.10^{**}$ & 751 (194) \\
\hline
$\text{Germanic}$ & 87.86 & 86.58 & 1.28 & 1.32 & 328 (101) \\
\hline
$\text{Other Indo-Euro}$ & 85.20 & 84.52 & 0.68 & 0.29 & 110 (27) \\
\hline
$\text{Ural-Altaic}$ & 91.72 & 85.22 & 6.50 & $3.11^{**}$ & 37 (9) \\
\hline
\end{tabular}
}
\caption{Paired t-test by language family. 
The difference is calculated by subtracting the pre-ChatGPT release $score$ from the post-release $score$.
The superscript $**$ indicates that this difference is significant at the 5\% level.
The column "Observation" displays the number of papers, with the number 
of authors shown in the parentheses.
}\label{Tab:Paired}
\end{table}

\section{Difference-in-Differences analysis}
\label{sec:ecoana}

\subsection{Review of method}

In the Welch's t-test and paired t-test above, we compare the 
difference in $score$ before and after the release of ChatGPT. 
However, this difference cannot be solely attributed to the 
release of ChatGPT, as several factors could influence writing 
style changes during this period. For instance, an author's 
English writing skills may naturally improve over time, or changes 
in co-authors, who may have varying levels of English proficiency, 
could also contribute to the observed differences. It is crucial
 to account for these factors to determine whether the change in 
 writing style is genuinely due to the adoption of ChatGPT.

To isolate the impact of ChatGPT on English writing style, 
we employ the Difference-in-Differences (DID) method, a causal 
inference technique widely used in econometrics and quantitative 
social science research. The DID approach helps mitigate the 
effects of extraneous factors and unobserved confounders, 
enabling us to assess whether the significant increase in writing quality 
is indeed attributable to ChatGPT.

We use native English speakers as the control group and 
non-native English speakers as the treatment group. First, 
we select an annual data frequency and calculate the means 
of the variables over one year to obtain annualized sample 
values for each author. After this averaging process, the total 
sample size is reduced to 1,331, resulting in an unbalanced 
panel dataset from 2020 to 2023.
In the DID method, the relationship between variables can be expressed as
\be\label{eq:did}
Y_{it} = \alpha + \beta_1 T_i  P_t + \sum_C \beta_C 
C_{it} + u_t  + v_i + e_{it},
\ee
where $Y_{it}$ denotes a dependent variable that describes 
the writing style of an abstract, such as $score$ (the most important one), 
${read\_score}$, $unique$, $rare$ and $issues$.
Here, $i$ denotes the author's ID, and $t$ denotes time.
The intercept, $\alpha$, is set to the baseline average for the 
control group before the release of ChatGPT. 
$T_i=1$, or 0 indicates the treatment or control group, respectively, while
$P_t$ indicates whether the paper was written before ($P_t=0$) or after ($P_t=1$)
the release of ChatGPT. Crucially, $\beta_1$ is the DID estimator 
that captures the treatment effect. In other words, 
$\beta_1$ isolates the effect of ChatGPT’s release by accounting 
for both preexisting differences between the groups and the overall 
time trend affecting both groups.

The control variables are represented by $C_{it}$, which include
$others\_Eng$, $authors$, $stat$, $dis$, ${if\_pub}$, $characters$, 
$words$, $sentences$, ${word\_len}$, and ${sentence\_len}$.
Descriptive statistics for these variables can be found in Tab.~\ref{Tab:Basic}.
Additionally, $papers$ denotes the number of papers authored by $i$ in year $t$,
reflecting the author’s writing activity.
Each control variable may influence the dependent variable, and 
their effects are measured by the coefficients $\beta_C$.

Correlation analysis shows that $characters$, $words$ and 
$sentences$ are highly correlated, with a correlation coefficient of $0.82$ between
$words$ and $sentences$, and $0.98$ between $words$ and $characters$. 
Therefore, in the subsequent DID analysis, we include 
$words$ as a control variable, while excluding $characters$ and $sentences$ to avoid multicollinearity.

$u_t$ captures any year-specific effects that could otherwise confound our results, 
while $v_i$ controls for time-invariant characteristics specific to each author.
In our DID analysis, we consistently consider time and author fixed effects. Finally, 
$e_{it}$ represents the random error term.

\subsection{Results of DID analysis}

\begin{table}[tbp]
\resizebox{0.9\linewidth}{!}{
\tiny
\renewcommand\arraystretch{1.0}
\begin{tabular}{| c | c | c | c | c |}
\hline
 & $score$ & $score$ & $score$ & $score$ \\
\hline
$T_i P_t $ & $1.05^{***}$ & $1.04^{***} $ & $0.71^{**}$ & $0.81^{**}$ \\
\hline
 & (4.64) & (4.54) & (2.73) & (3.13) \\
\hline
${others\_Eng}$ &  & $0.73^*$ & $0.82^*$ & $1.09^{**}$ \\
\hline
 &  & (1.96) & (2.19) & (2.54) \\
\hline
${authors}$ &  & 0.09 & 0.18 & 0.16 \\
\hline
 &  & (0.55) & (0.85) & (0.69) \\
\hline
${stat}$ &  & -0.67 & -0.80 & -0.72 \\
\hline
 &  & (-0.39) & (-0.41) & (-0.33) \\
\hline
${dis}$ &  & -0.69 & -0.61 & -0.69 \\
\hline
 &  & (-0.79) & (-0.94) & (-0.96) \\
\hline
${if\_pub}$ &  & 0.21 & 0.18 & 0.35 \\
\hline
 &  & (0.48) & (0.45) & (1.00) \\
 \hline
${papers}$ &  & -0.03 & -0.03 & -0.03 \\
\hline
 &  & (-0.09) & (-0.08) & (-0.10) \\
  \hline
${words}$ &  &  & $-0.03^{**}$ & $-0.03^{***}$ \\
  \hline
 &  &  & $(-3.66) $ & $(-4.00) $ \\
   \hline
${word\_len}$ &  &  &  & 0.46 \\
    \hline
 &  &  &  & (0.74) \\
     \hline
${sentence\_len}$ &  &  &  & $0.17^{**}$ \\
   \hline
 &  &  &  & (2.81) \\
    \hline
$\alpha$ & $86.75^{***}$ & $86.89^{***}$ & $91.31^{***}$ & $84.51^{***}$ \\
    \hline
 & (1544.53) & (51.87) & (31.89) & (13.89) \\
     \hline
Adjusted $R^2 $ & 0.14 & 0.13 & 0.15 & 0.16 \\
\hline
Observations  & 1331   & 1331 & 1331 & 1331  \\
         \hline
\end{tabular}
}
\caption{DID analysis of the dependent variable $score$.
The t-values, shown in parentheses, indicate the robustness of the results. 
The superscript ${*\! *\! *}$ denotes statistical significance at the 1\% level, 
while $**$ and $*$ denote significance at the 5\% and 10\% levels, respectively. 
The analysis includes 459 observations in the English-speaking group
and 872 observations in the non-English-speaking group.} 
\label{Tab:DIDscore}
\end{table}

Table~\ref{Tab:DIDscore} presents the results of DID analysis.
The adjusted $R^2$ is a refined measure of goodness-of-fit for our model~\eqref{eq:did}, 
indicating the percentage of variance in the dependent variable that is explained by the independent variables, 
adjusted for the number of predictors.
In Tab.~\ref{Tab:DIDscore}, the adjusted $R^2$ values range from 0.13 to 0.16, 
which is acceptable for our DID analysis.

% Adjusted R-squared is a corrected goodness-of-fit (model accuracy) measure for linear models. 
% It identifies the percentage of variance in the target field that is explained by the input or inputs.

Different columns in Tab.~\ref{Tab:DIDscore} correspond to various choices of the set of $C_{it}$. 
In the first column, no control variables are included, 
while in the fourth column, all nine control variables are incorporated. 
The estimated coefficients $\beta$ are presented in each row. 
For example,
the value of $\beta_1$ is provided in the row labeled "$T_iP_t$", 
and the coefficient $\beta_{C}$ for the control variable $others\_Eng$ is shown in the row labeled "$others\_Eng$".
Additionally, the t-value for each estimate is displayed in the parentheses directly below the corresponding estimated coefficient. 
The t-value indicates the statistical significance of the estimation.

The DID analysis reveals that after the release of ChatGPT, 
the treatment group (non-English speakers), 
shows a significant improvement in their $score$ at the 1\% level, 
with $\beta_1= 1.05$ when no control variable are included. 

According to model~\eqref{eq:did}, a positive  $\beta_1$ indicates an improvement in the dependent variable due to ChatGPT's release, 
while a negative  $\beta_1$ would suggest a reduction. 
% Note that $\beta_1>0$ ($\beta_1<0$) means an improvement (reduction) of dependent variable impacted by the release of ChatGPT, 
Columns 2 through 4 present the $\beta$ coefficient after controlling for various sets of variables.
In all cases, the release of ChatGPT has a statistically significant positive impact on the writing skills of the non-native authors, 
as evidenced by the consistently positive $\beta_1$. 
For example, 
after controlling for all variables, 
$\beta_1$ is estimated at 0.81, which is significant at the 5\% level (see the 4th column). 
Therefore, compared to native English speakers,
ChatGPT has significantly helped non-native speakers improve their writing quality,
suggesting widespread adoption of ChatGPT among non-native authors in scientific writing.

Among the control variables, the variable ${others\_Eng}$ 
has a positively significant effect on $score$, with a corresponding
coefficient of $\beta_C= 1.09$ in the 4th column.
This indicates that if a paper has English-native coauthors (apart from the first author), the writing quality tends to improve,
which is a reasonable outcome. 
Additionally, Tab.~\ref{Tab:DIDscore} shows
that the coefficient for $words$ is $-0.03<0$, which is significant at the 1\% level in the fourth column.
This suggests that a higher word count in the abstract is associated with lower writing quality.
Conversely, writing quality improves with longer sentences, 
as reflected by the positive coefficient of $0.17$ in the 4th column.

\begin{table}[tbp]
\centering
\renewcommand\arraystretch{1.0}
\resizebox{0.85\linewidth}{!}{
\begin{tabular}{| c  | c | c |  c | c | c | c | c | c |}
  \hline
                   &$read\_score$  & $unique$ &  $rare$ & $issues$ \\
\hline
$T_i P_t$    & $1.08^{***}$  & $0.01^{**}$ &  $-0.01^*$ & $-0.99^{***}$  \\
  \hline
& (4.03) & (3.57) & (-2.18) & (-5.93)  \\
 \hline
 $others\_Eng$     & 0.32    & 0.002 & $0.01^{***}$  & 0.03  \\
  \hline
 & (0.27)  & (0.26) & (5.22)  & (0.16)  \\
  \hline
 $authors$   & 0.26   & $-0.004^*$ & 0.002  & 0.07  \\
   \hline
 & (1.56)   & (-2.25) & (1.64)  & (1.49)  \\
    \hline
 $stat$   & 1.30   & 0.01 & 0.002  & 0.30  \\
    \hline
  & (0.69)   & (1.03) & (0.15)  & (0.38)  \\
      \hline
 $dis$   & 0.18   & -0.004 & -0.01  & 0.58  \\
     \hline
   & (0.09)   & (-0.40) & (-1.13)  & (0.88)  \\
    \hline
 $if\_pub$    & 1.02   & $-0.01^{**}$  & $-0.01^{**}$  & -0.12  \\
     \hline
   & (0.71)   & (-2.57)  & (-2.88)  & (-0.49)  \\
      \hline
 $papers$    & 0.10   & $-0.002$ & $0.0001$  & 0.04  \\
      \hline
   & (0.38)   & (-1.42) & (0.16)  & (0.38)  \\
      \hline
 $\alpha$   & $14.39^{***}$  & $0.62^{***}$ &   $0.36^{***}$ &  $8.02^{***}$  \\
       \hline
 & (6.60)  & (43.08) & (30.07) & (8.15) \\
\hline
Adjusted $R^2$   & 0.36   & 0.27 & 0.19  & 0.21  \\
         \hline
\end{tabular}
}
\caption{DID analysis of the dependent variables: $read\_score$, $unique$, $rare$, and $issues$.
The number of observations for each column is 1331.}\label{Tab:DIDother}
\end{table}

We further analyze other dependent variables, such as $read\_score$, 
$unique$, $rare$, and $issues$, which all characterize
the writing style of an abstract. 
The results are presented in Tab.~\ref{Tab:DIDother}. 
The readability of abstracts by non-native speakers
has significantly improved following the release of ChatGPT, 
with a coefficient $\beta_1$ of 1.08, significant at the 1\% level.
The adjusted $R^2$ value is 0.36, indicating that our model~\eqref{eq:did}
and choice of control variables explains 36\% of the variation in $read\_score$.
% or: explains $read\_score$ variation at 36\% level. 
% Adjusted R-squared is a corrected goodness-of-fit (model accuracy) measure for linear models. 
% It identifies the percentage of variance in the target field that is explained by the input or inputs.

For the dependent variable $unique$, we find that after the release of ChatGPT, 
the number of unique words used by 
non-native speakers significantly increased, with a coefficient $\beta_1$ of 0.01.
This suggests that with the assistance of ChatGPT, non-native speakers 
tend to utilize a greater variety of unique words. 
In contrast,
the use of rare words decreased after ChatGPT's release, with a coefficient of -0.01.
This may be because ChatGPT tends to avoid rare words in its generated content. 
Additionally, for the variable $issues$, we observe a significant reduction in writing issues among non-native speakers after ChatGPT's release, 
with a coefficient of -0.99 at the 1\% significance level.
This indicates that ChatGPT is indeed effective in correcting writing errors.

\begin{table}[tbp]
\renewcommand\arraystretch{1.0}
\resizebox{0.85\linewidth}{!}{
\begin{tabular}{| c  | c | c |  c | c | c | c | c | c |}
  \hline
  & (1) score  & (2) score & (3) score & (4) score  \\
\hline
 $T_iP_t$  & $0.87$  & $0.81^{**}$  & $1.45$ & $6.86^*$  \\
  \hline
 & (0.85)  & (2.91) & (0.61) & (3.42)  \\
 \hline
 $others\_Eng$    & 0.65   & $1.16^{**}$ & $1.41$  & $1.32^{**}$ \\
  \hline
   & (0.58)   & (3.44)  & (1.75)   & (5.65)  \\
  \hline
 $authors$    & -0.14    & $-0.01$  & $-0.12$   & -0.19  \\
   \hline
   & (-0.63)    & (-0.04)  & (-0.61)  & (-1.83)  \\
    \hline
 $stat$   & $1.54^*$  & -0.91 & 0.43  & $1.83^*$  \\
    \hline
   & (2.35)    & (-0.48) & (0.27)   & (3.39)  \\
      \hline
 $dis$   & -1.02  & -1.78 & -1.65  & $-2.29^{**}$ \\
     \hline
   & (-0.64)   & (-1.53)  & (-1.28)   & (-7.62)  \\
    \hline
 $if\_pub$     & 0.01    & $-0.20$  & $0.65$  & 0.24  \\
     \hline
   & (0.01)   & (-0.87)  & (0.79)   & (0.43)  \\
      \hline
 $papers$    & 0.37   & $-0.27$ & $0.37$  & 0.14  \\
      \hline
   & (1.70)   & (-0.84)  & (1.49)  & (1.36)  \\
      \hline
 $\alpha$   & $86.06^{***}$   & $88.40^{***}$ & $85.96^{***}$ & $86.24^{***}$  \\
       \hline
 & (176.81) & (56.89)& (149.03) & (113.21) \\
\hline
Adjusted $R^2$   & 0.13   & 0.15 
 & 0.10 & 0.17  \\
          \hline
Observations  & 709   & 976 & 537 & 486  \\
         \hline
\end{tabular}
}
\caption{DID analysis of $score$ for the effect of ChatGPT across different language families.
In all four columns, the 459 observations in the English-speaking group serve as the control group.
Column (1) takes 250 papers from the Germanic group as the treatment group. 
Column (2) focuses on 517 papers from the Latin group as the treatment group. 
Column (3) includes 78 papers from other Indo-Euro language families as the treatment group.
Finally, column (4) analyze 27 papers from the Ural-Altaic group as the treatment group.}
\label{Tab:DIDlangage}
\end{table}
459 observations in the English-speaking group
and 872 observations in the non-English-speaking group

To further distinguish the impact of ChatGPT's release across different 
language families, we conduct a detailed DID analysis by language family
(see Tab.~\ref{Tab:DIDlangage}).
For this analysis, we categorize the authors by their language
family, selecting one group (excluding the English group) as the treatment group and using the 459 papers from the English group as the control group.
As a result,
the sample size of each analysis is reduced to fewer than $1331$, 
with the exact number of observations indicated in the last row labeled "Observations".

In column~(1) of Tab.~\ref{Tab:DIDlangage}, we compare
the impact of ChatGPT on the writing quality of English speakers (control group)
and Germanic speakers (treatment group, excluding English speakers).
The results indicates that the release of ChatGPT does not lead to a statistically difference in writing quality between English and Germanic groups.
The coefficient $\beta_1$ is 0.87, which is not significant, even at the 10\% level.
This suggests that Germanic speakers do not experience a significant improvement in their writing skills 
compared to English native speakers.
A more probable reason could be that, since English is a sub-branch of the Germanic language
and shares similarities with it, Germanic speakers may not feel a strong need to rely on ChatGPT.
Similarly, in column~(3), we observe that the speakers from other Indo-Euro language families
do not show a significant improvement in their writing skills, with $\beta_1$ estimated at 1.45
(also not significant at the 10\% level).

In contrast, for the Latin group (see column~(2) of Tab.~\ref{Tab:DIDlangage}),
we observe a significant improvement following the release of ChatGPT, 
compared to the English group. 
The coefficient $\beta_1$ is 0.81, which is significant at the 5\% level. 
A similar significant improvement is also noted for
the Ural-Altaic group (see column~(4) of Tab.~\ref{Tab:DIDlangage}).  

Therefore, our DID analysis indicates that the release of 
ChatGPT significantly enhances English writing skills and readability for non-native English speakers. 
It increases their usage of unique words,
reduces the use of rare words, and decrease writing errors. 
These effects are particularly pronounced for authors from the Latin and Ural-Altaic language families, but not significant
for authors from the Germanic or and Indo-Euro Language families.

\subsection{Robust test}

\begin{table}[tbp]
\renewcommand\arraystretch{1.0}
\resizebox{0.88\linewidth}{!}{
\begin{tabular}{| c  | c | c |  c | c |}
  \hline
 & (1) $score$ & (2) $score$  & (3) $score$ & (4) $score$ \\
\hline
$T_iP_t$   & $1.23^{***}$ & $0.97^{***}$ &  $0.76^{*}$  &  $0.71^{*}$ \\
  \hline
  & (4.94) & (3.79) &  (2.36)   &  (2.13) \\
 \hline
 $others\_Eng$   &  & $1.30$ &     &  $0.75^{**}$ \\
  \hline
                    &  & (1.89)  &    &  (2.67) \\
  \hline
 $authors$   &  & 0.16     &    &  0.24   \\
   \hline
                    &  & (0.51)  &    &   (1.40) \\
    \hline
 $stat$         &  & 0.26     &    &   -0.74  \\
    \hline
                    &  & (0.15)  &     & (-0.32) \\
      \hline
 $dis$          &  & -0.63   &      & -1.02   \\
     \hline
                   &  & (-0.93) &       & (-1.56)\\
    \hline
 $if\_pub$   &  & 0.07     &       & $1.00^*$\\
     \hline
                   &  & (0.24)  &        &(2.07)  \\
      \hline
 $papers$   &  & -0.36  &          &-0.12  \\
      \hline
                   &  & (-1.55)&          &(-0.43)  \\
  \hline
   $words$   &  & $-0.03^{***}$ &     &  $-0.03^{**}$   \\
   \hline
                   & & (-6.33) &          & (-3.64) \\
 \hline
 $word\_len$ &  & $0.80$ &       & $1.21^*$  \\ 
    \hline
                     & & (1.23)  &         &    (2.15) \\ 
      \hline
 $sentence\_len$  &  &  $0.18^{***}$  &        &  $0.18^{***}$\\  
    \hline
                            & & (4.21)  &           & (4.05) \\
   \hline
 $\alpha$   & $86.75^{***}$ & $82.35^{***}$ &   $87.09^{***}$  & $79.87^{***}$ \\
       \hline
  & (1608.01)  & (15.16) &  (1088.33)   &  (13.71) \\
\hline
Adjusted $R^2$ & 0.14 & 0.18 &   0.13    & 0.16   \\
          \hline
Observations  & 1641 & 1641  & 1173  & 1173 \\
         \hline
\end{tabular}
}
\caption{Robust test of DID. The columns~(1) and~(2)
display the quarterly-frequency comparison between
English and Non-English groups. The columns~(3) 
and~(4) display the yearly-frequency comparison
after deleting the samples in December 2022.
}\label{Tab:Robust1}
\end{table}

The above DID analysis is based on annualized data. To test the 
robustness of our results against different data frequencies, we 
also perform a DID analysis using quarterly data. Following a similar 
procedure as described earlier, we calculate the average values 
of each variable within each quarter for each author.
This process yields a total of 1,641 samples 
from the first quarter of 2020 to the second quarter of 2023, with 
559 samples in the English group and 1,082 samples 
in the Non-English group. We then use this quarterly data to conduct the DID analysis.

The results are presented in the columns~(1) and~(2) of Tab.~\ref{Tab:Robust1}.
In column~(2), we observe a $\beta_1$ value of $0.97$, which is positively significant 
at the 1\% level, after considering all the control variables. 
Column~(1) presents the coefficient $\beta_1$ without considering
any control variables. When comparing Tab.~\ref{Tab:Robust1}
with Tab.~\ref{Tab:DIDscore}, we find no qualitative differences 
between the results, indicating that our previous findings are robust 
regardless of whether we use yearly or quarterly data.

Additionally, we explore the possibility that the impact of ChatGPT's 
release might have been delayed due to the community's gradual 
adoption of such a revolutionary tool. To test this, we exclude the 
sample papers of December 2022 and redo the yearly-frequency DID analysis.
After this exclusion, the total sample size is reduced to 1,173, 
with 399 samples in the English group and 774 in the Non-English group. 
The results of this analysis are presented in columns~(3) and~(4) of Tab.~\ref{Tab:Robust1}.
In column (4), we find that the writing quality of non-English speakers 
improves at the 10\% significance level after ChatGPT's 
release, with a $\beta_1$ value of 0.71. This result qualitatively aligns 
with our previous findings. However, compared to the earlier results 
that included December 2022 samples (see the fourth column of Tab.~\ref{Tab:DIDscore}), 
the $\beta_1$ value is lower, and the significance level is also reduced. 
This suggests that ChatGPT had an immediate impact on non-English speakers,
who were eager to adopt the tool right after its release.

\section{Discussion}
\label{sec:summ}

In this paper, we systematically estimate the impact of ChatGPT's 
release on the writing style of condensed matter academic papers, 
using arXiv as the data source and "Grammarly" to quantify the 
writing style of abstracts. To ensure robust conclusions, we employ 
various statistical tools, ranging from Welch's t-test and paired 
t-test to the more advanced Difference-in-Differences (DID) method.

The t-test results reveal that writing quality significantly improves 
after the release of ChatGPT, while no significant differences are 
observed in the years leading up to its release. Specifically, 
for non-native English speakers, the improvement in writing quality 
is significant at the 5\% level, particularly within the Latin and 
Ural-Altaic language groups. 
By using the DID method to control for unobserved heterogeneity and other confounders, 
we find statistically significant improvements in writing quality for authors in the Latin 
and Ural-Altaic language groups, but not for those in the Germanic or other Indo-Euro language groups.
These findings suggest widespread adoption of 
ChatGPT among authors in the Latin and Ural-Altaic language families.

We confirm that the changes in writing style among Non-English speakers can indeed be attributed to the 
release of ChatGPT. Our DID analysis shows that ChatGPT not only enhances 
the quality and readability of papers by non-native English speakers 
but also alters their writing style. After ChatGPT's release, we 
observe an increase in the use of unique words and a decrease 
in the use of rare words, along with a reduction in writing issues.

However, our research has limitations. We focus solely on the changes 
in writing style due to ChatGPT adoption, without considering other 
large language models. Moreover, our study is based on short-term 
data, from January 2020 (the start of the COVID-19 pandemic) to 
May 2023, and does not include papers published after June 2023. 
Future studies should cover a broader time period. Additionally, 
our findings rely on "Grammarly" for scoring, which may not fully 
capture the nuances of academic writing quality. Exploring alternative 
scoring tools in future research would be beneficial.

It is important to clarify that we do not claim any researchers are 
using ChatGPT to write entire papers. Our analysis focuses on 
abstracts, where we find significant improvements in English 
quality due to ChatGPT's release. For example, using ChatGPT 
solely as a copyediting tool could lead to such improvements. 
It is crucial to differentiate between using ChatGPT for copyediting 
and using it to write entire papers. Some critics worry about potential 
plagiarism, lower quality, or increased errors when using ChatGPT 
in academic writing. There are also concerns that ChatGPT's quality 
may degrade if it is trained on AI-generated content.

Our study highlights the widespread use of ChatGPT in scientific 
writing, particularly within the condensed matter community. 
This trend appears difficult to avoid. More important, ChatGPT 
enhances the writing quality of non-native English authors and 
reduces grammatical errors in their work. As a result, ChatGPT is 
influencing the writing style within the condensed matter community, 
making it unlikely for anyone to remain unaffected. Journal editors 
now face the challenge of creating guidelines to distinguish between 
simple copyediting and more extensive use of ChatGPT, which is an urgent task.

\begin{acknowledgments}
S. Xu is supported by NSSFC under Grant No.~19BJL066.
 P. Wang is supported by NSFC under Grant No.~11774315, and the Junior Associates
program of the Abdus Salam International Center for Theoretical Physics.

\end{acknowledgments}

\end{document}